\documentclass[11pt]{article}
\usepackage[]{acl}
\usepackage{times}
\usepackage{latexsym}
\usepackage[T1]{fontenc}
\usepackage[utf8]{inputenc}
\usepackage{microtype}
\usepackage{inconsolata}
\usepackage{graphicx}
\usepackage{outlines}
\usepackage{booktabs}
\usepackage{todonotes}

\usepackage{pifont} 
\usepackage{multirow}
\usepackage{tabularx}
\usepackage[table]{xcolor}
\usepackage{listings}
\usepackage{tcolorbox}
\usepackage{multicol}

\newcounter{stepcounter}
\refstepcounter{stepcounter}

\newtcolorbox[auto counter]{mybox}[1]{%
colback=cyan!5!white,
colframe=cyan!75!black,
fonttitle=\bfseries,
title=Result~\thetcbcounter: #1}

\usepackage{subcaption}
\usepackage{amsmath}
\definecolor{myrowcolor}{HTML}{F2F2F2}
\definecolor{keywordcolor}{HTML}{770489}
\lstdefinelanguage{SPARQL}{
  keywords={SELECT, WHERE, FILTER, OPTIONAL, PREFIX, LIMIT, DISTINCT, GROUP, ORDER, BY, DESC, COUNT},
  sensitive=true,
  morecomment=[l]{\#},
}

\usepackage{adjustbox}

\begin{document}

\title{Auditing a KB Elicitation of Frontier LLM Knowledge: \\A Multi-dimensional Analysis of GPTKB v1.5}
\author{
\textbf{Shrestha Ghosh$^1$},
\textbf{Luca Giordano$^2$},
\textbf{Yujia Hu$^2$},
\textbf{Tuan-Phong Nguyen$^3$},
\textbf{Simon Razniewski$^2$}\\
\\
$^1$ University of Tübingen \ \
$^2$ ScaDS.AI \& TU Dresden \\
$^3$ VNU University of Engineering and Technology\\
\small\texttt{shrestha.ghosh@uni-tuebingen.de} \ \ \small\texttt{tuanphong@vnu.edu.vn}\\ 
\small\texttt{\{simon.razniewski,yujia.hu,luca.giordano\}@tu-dresden.de} \ \ 
}

\maketitle

\begin{abstract}
LLMs are remarkable artifacts that have revolutionized a range of knowledge-intensive tasks. A significant contributor is their factual knowledge, which, to date, remains poorly understood, and is usually analyzed from biased samples.
In this paper, we provide a framework and the results of a multi-dimensional analysis of GPTKB v1.5 \cite{GPTKB15}, a recursively elicited Knowledge Base (KB) of 100 million facts (or beliefs) of a frontier LLM, namely, GPT-4.1.
Given the scale of the elicited facts, we provide a multi-dimensional approach to qualitatively and quantitatively analyze these facts as opposed to the mainstream fact completion benchmarks, which are prone to availability bias. 
We find that the models' factual knowledge differs quite significantly from established knowledge bases, and that its accuracy is significantly lower than indicated by previous benchmarks. We also find that inconsistency, ambiguity and hallucinations are major issues, shedding light on future research opportunities in neuro-symbolic AI concerning extraction, consolidation and verification of factual LLM knowledge. 
\end{abstract}

\section{Introduction}

\paragraph{Motivation} Large Language Models (LLMs) are remarkable artifacts that have demonstrated impressive performance across a range of tasks such as coding, math (Claude 3.5 Sonnet 92\% in HumanEval and 91.6\% in MGSM~\cite{anthropic2024claude35sonnet}),
or general language understanding (GPT-4.1 90.2\% on MMLU~\cite{openai2025gpt41}). A major driver of their success is their internalized knowledge, however, little is understood about what knowledge they have actually internalized, what the accuracy of their knowledge is, and what biases this exhibits.

It may be tempting to prompt the models directly, however, as shown in Table~\ref{fig:teaser} (bottom left), such questions are usually answered in an unhelpful or incorrect way. Recently, the construction of epistemic twins from LLMs has been proposed as a way to comprehend LLM knowledge \cite{nguyen2022materialized,cohen2023crawling,hu2025enabling, razniewski2026epistemic}. Unlike sample-based studies \cite{petroni2019language,veseli2023evaluating,sun2023headtotail,amirizaniani2024auditLLM}, epistemic twins provide a view beyond reconstructing existing data via persistent knowledge base (KB) resource for repeated investigation \cite{hu2025enabling}. 

We answer four research questions to determine whether such a massive-scaled LLM-mined KBs serve as reliable and under what conditions. \textbf{RQ1.} How does the scale and coverage of LLM-mined knowledge compare to Wikimedia-related KBs? 
\textbf{RQ2.} What systematic biases does LLM-mined knowledge exhibit relative to real-world distributions?
\textbf{RQ3.} How does factual accuracy change with long-tailed entities?
\textbf{RQ4.} How consistent and temporally coherent is the mined KB?


\begin{table}[t]
\resizebox{1\columnwidth}{!}{
\centering
\large
\begin{tabular}{@{}p{6cm}@{\qquad} p{6cm}@{}}
\toprule
\textbf{Direct   LLM querying} & \textbf{LLM-mined knowledge analysis}   \\ 
\midrule
\textcolor{red}{\ding{55}} Lacks aggregate access to \qquad internalized knowledge & \textcolor{green!60!black}{\ding{51}} Accesses millions of extracted records   \\
\textcolor{red}{\ding{55}} Gives uninformative and wrong answers & \textcolor{green!60!black}{\ding{51}} Provides quantitative insights   \\ \midrule

\textbf{ChatGPT interface} & \textbf{Mined knowledge \mbox{(GPTKB v1.5)}}   \\ \midrule
\multicolumn{2}{l}{\qquad\qquad \textit{\textbf{How many individual persons do you know?}}}   \\
{\small I don’t ``know'' any individual persons [..]. [..] millions to billions, depending on [..].} \textcolor{red}{$\rightarrow$uninformative} & \qquad 1,200,692   \\[9pt]
\multicolumn{2}{l}{\qquad \qquad \textit{\textbf{Do you know more males or more females?}}}   \\ 
{\small I don't have a built-in count of individuals by gender. However, in the publicly available sources I draw from [..]. So [..] I would ``know'' more males than females.} \textcolor{red}{$\rightarrow$false} & \qquad 1 : 1.16 \\[9pt]
\multicolumn{2}{l}{\qquad \qquad \textit{\textbf{Do you know more about Chinese, or Canadian entities?}}}   \\
{\small I likely ``know'' more about Chinese entities than Canadian ones, simply because China [..] more [..] documented.} \textcolor{red}{$\rightarrow$false} & \qquad 1 : 3.29 \\ 
\bottomrule
\end{tabular}
}
\caption{Direct LLM querying, vs. large-mined knowledge analysis. ChatGPT conversation archived at {\small\url{https://chatgpt.com/share/68b6cd03-85e4-800d-81c8-afebc74ccc9d}.}}
\label{fig:teaser}
\end{table}

\paragraph{Contribution}
We provide a framework and the results of a multi-dimensional analysis of the knowledge of a frontier LLM, GPT-4.1, as surfaced via recursive elicitation and consolidation into GPTKB v1.5~\cite{GPTKB15}. We find:
\begin{enumerate}
    \item Frontier LLMs can contain a vast amount of factual knowledge, with quite different focus and biases than existing KBs.
    \item At overall 75\% accuracy, the accuracy of GPT's knowledge is much higher than previous text-extraction-based KBs, but also significantly lower than human-curated resources and results on popular LLM benchmarks.
    \item Inconsistency, ambiguity, and hallucinations are major issues, shedding light on the challenges and future research opportunities of extraction, consolidation and verification of factual LLM knowledge.
\end{enumerate}
Our analysis specifically uses GPTKB v1.5, an epistemic twin constructed for over \$14,000 from GPT-4.1, containing over 100M factual assertions for 6M entities, and there is no comparable project based on other LLMs available at the moment. While parts of our results are therefore specific to this LLM version, GPT-4.1 is an archetype of commercial frontier LLMs, and our analysis methodology generalizes beyond this model. Moreover, our way of analyzing closed-source model is especially curious as it provides an exemplary case study on how knowledge materialization can be used to reverse-engineer internals of closed-source models to reveal the knowledge/beliefs held by a frontier LLM (Table~\ref{fig:teaser} right). 
We use the dataset and code provided on the project page.\footnote{\url{https://gptkb.org}} The code for our multi-dimensional analysis is available at \url{https://github.com/Knowledge-aware-AI/gptkb-analysis}.

\section{Background}

\paragraph{Knowledge Probing of LLMs} \citet{petroni2019language} had the first intuition to consider LMs as knowledge bases and introduced the LAMA benchmark, which consists of cloze-style statements derived from existing \textit{(s, p, o)} triples and question–answer datasets. 
This was followed by multiple similar efforts to test LLM factuality on multilingual data~\cite{kassner-etal-2021-multilingual}, head-to-tail popularity~\cite{sun2023headtotail}, unbiased Wikidata samples~\cite{veseli2023evaluating} and unseen events with known entities~\cite{zheng2024reliable}.
However, these works on benchmark-based ``knowledge probing'', despite attempts to mitigate the issue with automatic prompt optimization (e.g. \citet{jiang2020how,shin-etal-2020-autoprompt,wu2024towards}) as pointed out by~\citet{hu2025enabling}, suffer from availability bias \cite{kahnemann}, i.e. they cannot reveal knowledge outside the benchmark’s scope.

\paragraph{LLM Knowledge Extraction and Epistemic Twins}
\citet{cohen2023crawling} propose LMCRAWL that extracts a knowledge graph of facts from a language model.
Given a seed entity, first its paraphrases are generated, then all possible relations and their respective paraphrases, and finally objects for all relations.
\citet{parovic-etal-2025-generating} propose the CoVe method to extract domain-specific KGs from LLMs with an LLM-generated schema and verification prompts. Both these methods are limited in their scope of knowledge extracted from LLMs. 
LMCRAWL generates less than 100 triples per seed entity in two hops and the CoVe KGs on books and landmarks contain tens of thousands of entities and hundreds of thousands of triples per domain. 
The idea has been lifted into the epistemic twin paradigm: The construction of comprehensive approximations of an LLM's knowledge in symbolic form \cite{razniewski2026epistemic}.

\paragraph{GPTKB v1.5} 
The GPTKB methodology \cite{hu2025enabling} combines a recursive knowledge elicitation process with a post-hoc knowledge consolidation phase.
Starting from a seed subject, the LLM is prompted to return knowledge about it in the form of triples. New named entities in these triple objects are identified via LLM-based named-entity recognition (NER) for further elicitation in a recursive BFS-based graph exploration process.
To address the redundancy and variance introduced during knowledge
elicitation, 
a greedy clustering algorithm is used to iteratively merge relations and classes into more frequent ones, given a sufficiently high label embedding similarity.
This approach at ``knowledge crawling'' mitigates the availability bias typical of probing-based benchmarks and enables knowledge analysis more flexible than memorization quantification approaches, which depend on access to training data and the LLM's ability to generate exact substrings of the training data \cite{carlini2023quantifying,carlini2023extracting}. 

\begin{table}[t]
\begin{center}
\resizebox{\columnwidth}{!}{%
\begin{tabular}{@{}lrr@{}}
\toprule
\textbf{Resource} & \textbf{\#entities} & \ \ \  \textbf{\#assertions} \\ \midrule
\textit{Wikimedia-related} \\
Wikidata \cite{wikidata} & 113M & 1.62B \\
Yago 4.5 \cite{suchanek2024yago} & 50M & 140M  \\
DBpedia \cite{dbpedia} & 3.8M & 75M \\ 
\midrule
\textit{Text-extracted} \\
NELL \cite{carlson2010toward} & 1.9M & 12M \\
ReVerb \cite{fader-etal-2011-identifying} & 2.3M & 15M \\ 
\midrule
\textit{Generative} \\
LMCRAWL \cite{cohen2023crawling} & ? & 69.8 $\pm$ 52.9 \\
CoVe \cite{parovic-etal-2025-generating} & 22K-65K & 53K-153K \\
GPTKB v1.5 \cite{GPTKB15} & 6.1M & 100M  \\ 
\midrule
\textit{LLM Factuality Benchmarks} \\
Head-to-Tail \cite{sun2023headtotail} & ? & 18K\\
LAMA Probe \cite{petroni2019language} & ? & 50K \\
WD-KNOWN \cite{veseli2023evaluating} & 2.9M & 3.9M \\ 
\bottomrule
\end{tabular}
}
\end{center}
\caption{Size comparison of knowledge resources and LLM factuality benchmarks. 
}
\label{tab:KB-sizes}
\end{table}

\section{Content Scale and Coverage}

In this section we address \textit{RQ1. ``How does the scale and coverage of LLM-mined knowledge compare to Wikimedia-related KBs''}?

\paragraph{Size} The very first observation is that there is a lot of knowledge in GPT, orders of magnitude more than tested for in any previous benchmark.
Table \ref{tab:KB-sizes} relates the size of GPTKB v1.5 to other prominent resources and benchmarks. We find that the size is on par with major existing KBs like DBpedia and Yago, as well as major text extraction projects. Only Wikidata is still an order of magnitude larger, although its biases are well documented, e.g., an undue weight on scholarly information (45.5M scholarly article entities vs. 77.3M other entities as of July, 2026) \cite{wikidata-wdqs-graph-split}.

\paragraph{Topical Analysis}
The most frequent classes and relations in GPTKB are shown in Table \ref{tab:gptkb_classes_relations}. \textit{person} and \textit{human} dominate the KB at 1.2M making up 19\% of all entities, followed by \textit{film} and \textit{company} at just 4\%. This lower-bound estimate of the number of persons the model has acquired data on is especially remarkable, as OpenAI is notoriously secretive about the amount of data seen or learned by their models, with their models consistently denying storing any data about individuals (see Tab.~\ref{fig:teaser}).~\footnote{Note that we do not claim violations of privacy here, as the data may be voluntarily published, and sourced from public resources like Wikidata. More on sensitive data below.}
The remaining classes concern media artifacts (film, book, song, album), geopolitical entities (city, village), organizations (company). This is in stark contrast to Wikidata, where the dominating class, scholarly articles, makes up 38\% of the KB, followed by humans (10\%) and astronomical objects (7\%).
The mandatory \textit{instanceOf} relation aside, the relation frequency show a more even distribution.
In contrast to Wikidata, GPTKB puts much more emphasis on so-called \textit{soft/subjective} relations that narrate achievements, like \textit{notableWork, notableFor}, which are among the most frequent.

\begin{table}[t]
\centering
\resizebox{\columnwidth}{!}{%
\begin{tabular}{l r @{\hskip 1cm} l r}
\toprule
\textbf{Class} & \textbf{Frequency} & \textbf{Relation} & \textbf{Frequency} \\
\midrule
person & 1,061,852 & instanceOf & 6,588,295 \\
human & 138,840 & locatedIn & 2,121,234 \\
film & 123,021 & occupation & 1,962,461 \\
company & 122,847 & country & 1,927,856 \\
book & 112,188 & notableWork & 1,500,584 \\
song & 103,824 & genre & 1,268,067 \\
fictional\_character\!\!\!\! & 93,779 & nationality & 1,236,491 \\
album & 81,500 & language & 1,224,199 \\
city & 80,861 & location & 1,072,795 \\
village & 78,753 & notableFor & 977,561 \\
\bottomrule
\end{tabular}%
}
\caption{Most frequent GPTKB classes and relations.}
\label{tab:gptkb_classes_relations}
\end{table}

\begin{table}[t]
\centering
\resizebox{\columnwidth}{!}{%
\begin{tabular}{lr@{\hskip 1cm}lr}
\toprule
\textbf{Domain} & \textbf{Fraction} & \textbf{Domain} & \textbf{Fraction}\\
\midrule
Science        & 0.2\% & Business & 7.0\% \\
Technology    & 4.0\%  & Health & 0.0\% \\
Engineering   & 1.3\% & Environment & 1.9\% \\
Mathematics  & 0.0\%  &  event & 0.8\% \\
Arts        & 23.8\% & unknown & 0.0\% \\
Humanities     & 60.0\%  & food & 1.1\% \\
Social Sciences       & 0.2\% & right & 0.0\% \\
\bottomrule
\end{tabular}
}
\caption{Entity proportion of first-order subbranches of the taxonomy, based on a random sample of 1000 entities.}
\label{tab:taxonomy-proportion}
\end{table}

\paragraph{Taxonomy}
The seed taxonomy generated by the LLM, without priors from the resource KB, exhibited an architectural bias towards STEM. It established \textit{Engineering}, \textit{Mathematics}, \textit{Technology} and \textit{Science} as peer categories to \textit{Humanities} and \textit{Arts}. In stark contrast, when this STEM-centric taxonomy was populated, the taxonomy is overwhelmingly concentrated in the Humanities and Arts branches (60.0\% and 23.8\%), while Technology (4.0\%), Engineering (1.3\%) and Science (0.2\%) contribute only a very small proportion (see Table~\ref{tab:taxonomy-proportion}). This uncovers a critical decoupling between the LLM's organizational logic and its knowledge content and demonstrates that the LLM's worldview is twofold. 

\begin{figure}[t]
     \centering
      \includegraphics[width=\columnwidth,]{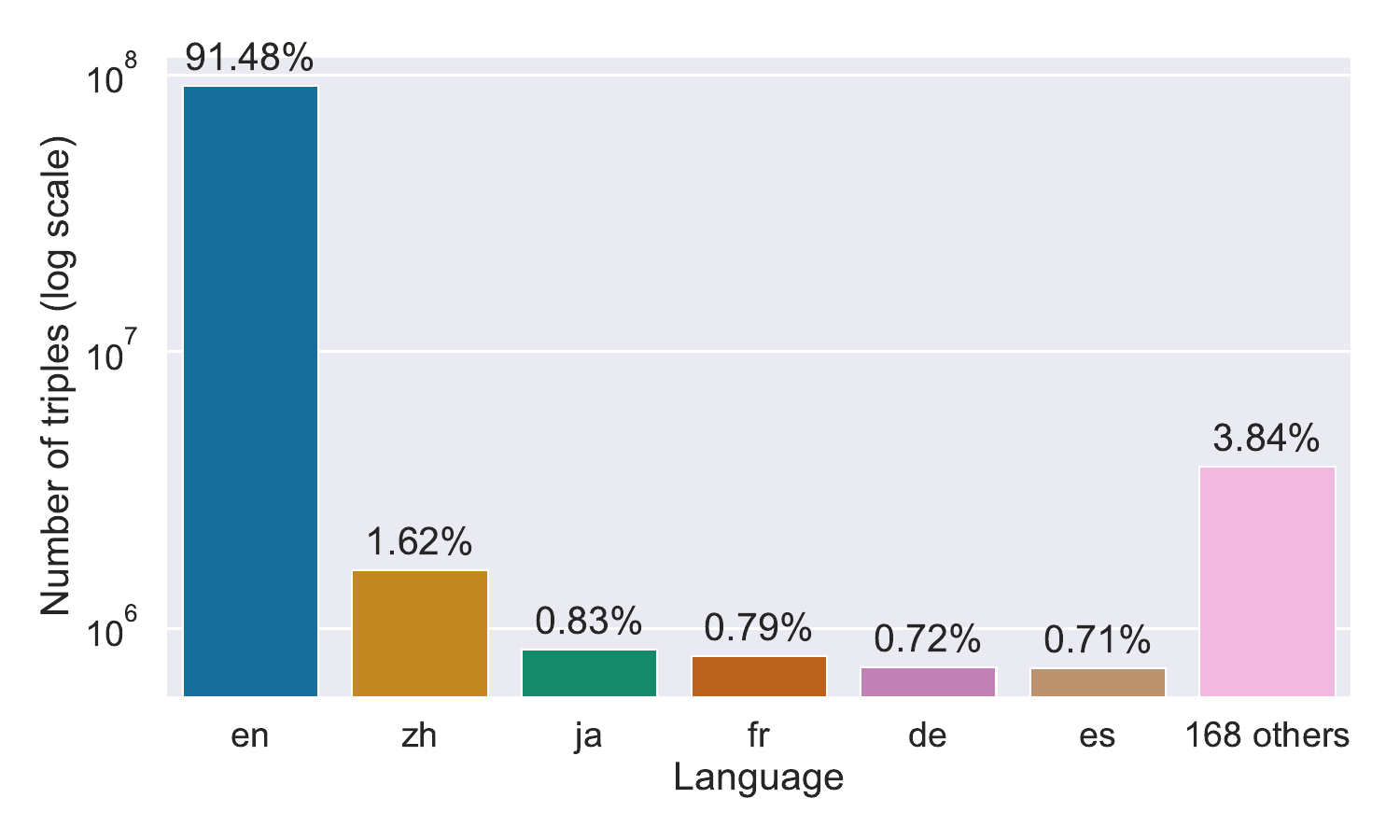}
     \caption{Language distribution in GPTKB.}
     \label{figure:languagedistribution}
 \end{figure}

\paragraph{Language} 
We performed language detection on the triples using fastText ~\cite{joulin2016bag} to analyze the linguistic composition of GPTKB. Figure~\ref{figure:languagedistribution} shows the distribution of the six most frequent languages, with the remaining languages aggregated.
Given that GPTKB was constructed using English-based prompts, the fact that English dominates at 91.5\% of the content is unsurprising.
Crucially, in our analysis of the remaining 9.5\% of the detected content, we found that non-English knowledge is largely self-contained. Only 22.6\% of all knowledge paths that start from an English entity and move to a non-English one return to an English-language statement. Particularly, Japanese, Chinese and Hebrew entities return to English in less than 10\% of the cases.
This evidence strongly indicates that much of the LLM's non-English knowledge is organized into its own densely interconnected, language-specific clusters. 
The long tail of 168 languages collectively accounts for 3.84\% of the KB's content. Such a substantial and diverse linguistic footprint, elicited from English-centric queries, suggests that the internal knowledge of the LLM contains a rich repository of multilingual content that remains largely unexplored.

\begin{table}[ht]
\centering
\small
\begin{tabular}{lr@{\qquad}lr}
\toprule
\textbf{Property} & \textbf{Count} & \textbf{Property} & \textbf{Count} \\
\midrule
logo        & 37,338 & mapType     & 2,432 \\
coverArt    & 33,717 & photo       & 2,253 \\
signature   & 32,443 & portrait    & 1,337 \\
coatOfArms  & 5,833 & range       &   581 \\
flag        & 3,424 & hasPoster   &   517 \\
\bottomrule
\end{tabular}
\caption{Top-10 predicates which come with objects being image file names.}
\label{tab:top-predicates-of-images}
\end{table}

\paragraph{Literals}
The are 42.9M triples with literal objects in GPTKB, making up 43\% of all triples.
\textit{website} and \textit{occupation} are the most frequent predicates with literals objects (frequency > 800K). This is followed by predicates denoting language \textit{language}, notability \textit{notableFor} and \textit{knownFor}, and date-related predicates such as \textit{birthDate} and \textit{deathDate}. 
Interestingly, we also found image filenames as literal objects occurring most frequently with predicates, such as, \textit{logo}, \textit{coverArt} and \textit{signature} (Table \ref{tab:top-predicates-of-images}).
This indicates that the LLM has memorized some image filenames, likely from resources such as Wikipedia or Wikidata.

\paragraph{Sensitive Data}
\label{sensitive_data}
We heuristically identified predicates in GPTKB on eight potentially sensitive data dimensions, namely, 
gender, race/ethnicity, religion, disability, sexual orientation, 
contact details, financial and military information.
Table \ref{tab:sensitive_predicates}
lists the identified predicates and the count of subjects associated with them.
Notably, personal identifiers appear only for public entities (e.g., address and phone number for organizations, net worth for celebrities). Entities who are children. i.e. who have a birth date less than ten years from the time of writing (August 2025) also do not raise any privacy concerns on any of our predefined dimensions.
Inspired by \citet{dodge2021documenting}, which documented an unexpected amount of text from US military websites in the C4 corpus, in GPTKB we observe few military-related predicates, which contain only public domain information.

\begin{table}[t]
    \centering
    \adjustbox{max width=\linewidth}{%
    \begin{tabular}{p{2cm}p{2.5cm}p{5cm}p{1.7cm}}
        \toprule
        \textbf{Dimension} & \textbf{RegEx Heuristic} & \textbf{Predicates} & \textbf{\# Unique Subjects} \\
        \midrule
        Gender & *gender* & gender
        & 270,694 \\ \cmidrule{2-4}
        
        Race & *race* & race & 395 \\ \cmidrule{2-4}
        
        Ethnicity & *ethnicity* & ethnicity 
        & 49,799 \\ \cmidrule{2-4}
        
        Religion & *religion* & religion 
        & 81,675 \\ \cmidrule{2-4}
        
        Disability & *disability* & disability
        & 1,273 \\ \cmidrule{2-4}
        
        Sexual orient. & *sexualorient* & sexualOrientation & 4,594 \\ \cmidrule{2-4}
        
        \multirow{3}{*}{Contact details} & *address* & - 
        & 0 \\ 
        & *phone* & phoneNumber 
        & 2 \\ 
        & *email* & 
        hasEmail 
        & 1 \\ \cmidrule{2-4}
        
        \multirow{2}{*}{Finan. info.}& *income* &-
        & 0 \\ 
        & *networth* & netWorth / netWorthAtDeath
        & 2,253 \\ \cmidrule{2-4}
        
        \multirow{2}{*}{Military info.} & *coordinates* & coordinates 
        & 55,265 \\ 
        & *planneddeploy* & 
        plannedDeploymentSite / plannedDeploymentYear / plannedDeploymentRegion & 3 \\ 
        & *baseloc* & 
        hasBaseLocation / majorUSBaseLocation / militaryBaseLocation & 5 \\ 
        \bottomrule
    \end{tabular}
    }
    \caption{Predicates related to potential sensitive information found in GPTKB and count of unique subjects. We filter for instanceOf: human/person for all the predicates except those related to military info.}
    \label{tab:sensitive_predicates}
\end{table}

Gender, race/ethnicity, religion, disability, and sexual orientation information exist mainly for individuals where these dimensions are contextually relevant for their notability. For example, the most frequent race/ethnicity, "African-American", occurred almost exclusively for notable activists, jazz musicians, victims of hate crimes, and politicians from the Reconstruction era and the Civil Rights Movement.
The same finding is valid for religion (e.g. notable bishops, rabbis, monks, priests, and other religious leaders), disability (e.g.
Paralympic athletes), and sexual orientation (e.g. activists and celebrities supporting LGBTQ rights).

\begin{mybox}{Content}
GPTKB departs from Wikidata through diverse classes and subjective relations. The duality of the broad taxonomy and concentrated class population exposes the twofold worldview of GPT-4.1. Non-English content is organized in densely-connected, language-specific clusters.
\end{mybox}

\begin{figure*}[t]
        \centering
        \includegraphics[width=0.85\textwidth]{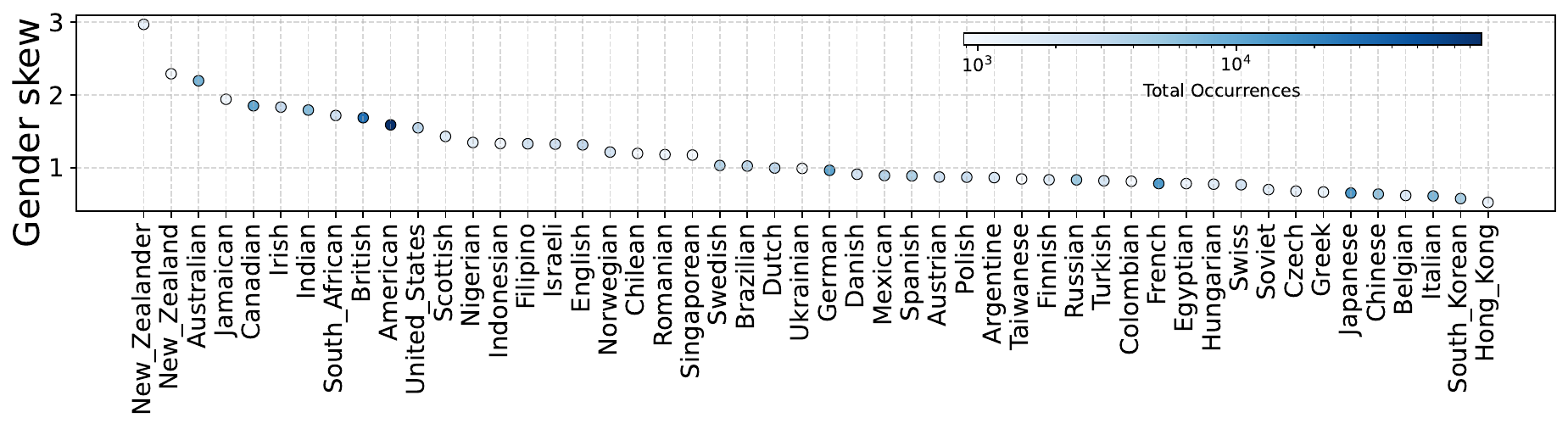}
        \caption{Female-to-male ratio (y-axis) in 50 most frequent nationalities (x-axis).
    }
    \label{fig:gender-ratio-by-nationality}
\end{figure*}
\begin{figure*}[htbp]
        \centering
        \includegraphics[width=0.85\textwidth]{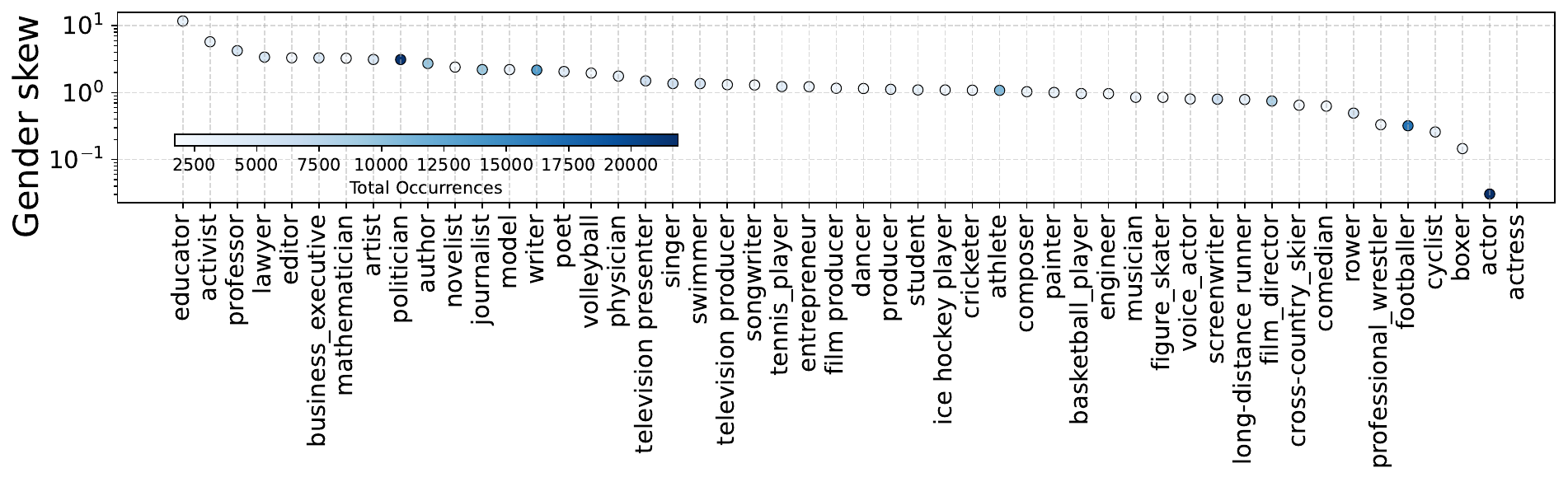}
        \caption{Female-to-male ratio (y-axis) in 50 most frequent professions (x-axis).
    }
        \label{fig:gender-ratio-by-profession}
\end{figure*}

\section{Biases}
In this section we address \textit{RQ2. ``What systematic biases does LLM-mined knowledge exhibit relative to real-world distributions?''}
\paragraph{Gender Bias} 
\textit{female} and \textit{male} are the dominant gender values in GPTKB, occurring in 196K and 168K triples respectively, with few other values, such as, \textit{non-binary} (589) and \textit{transgender woman} (413).
Notably, GPTKB has a female-to-male gender ratio of 1.16, i.e., \textit{female} is more frequent. 
The classes with the highest gender statements are \textit{person}, \textit{fictional\_character} and \textit{human}, with a gender ratio of 1.31, 0.68, and 0.88, respectively.
For comparison, in the real-world, this gender ratio, hovers close to but slightly below one \cite{ritchie2019gender}, and in Wikidata it is about 0.2 \cite{yunus2025exploring}.
It is notable that the gender is mentioned in just about 22\% of the entities from the classes \textit{person} and \textit{human}, reinforcing our previous observation about gender, among other personal identifiers, being mentioned only when contextually relevant.

\begin{table}[ht]
    \centering
    \resizebox{\columnwidth}{!}{
    \begin{tabular}{lrlr}
    \toprule
       \multicolumn{2}{c}{\textbf{No female gender}}  &  \multicolumn{2}{c}{\textbf{No male gender}}\\
       \cmidrule(lr){1-2} \cmidrule(lr){3-4}
       \textbf{Profession} & \textbf{Frequency} & \textbf{Profession} & \textbf{Frequency} \\
       \cmidrule(lr){1-2} \cmidrule(lr){3-4}
       sports shooter & 127 & gptkb:actress & 11248\\
cowboy & 74 & gptkb:businesswoman & 1307\\
gptkb:drag\_queen & 70 & noblewoman & 581\\
crime lord & 68 & gptkb:suffragist & 469\\
gptkb:hurler & 60 & gptkb:feminist & 323\\
gentleman & 55 & housewife & 269\\
gptkb:cryptocurrency & 48 & gptkb:beauty\_pageant & 226\\
drug dealer & 42 & waitress & 165\\
carpenter & 42 & homemaker & 164\\
gptkb:casino & 40 & gptkb:softball & 153\\
\bottomrule
    \end{tabular}
    }
    \caption{Top 10 gender-specific professions.}
    \label{tab:gendered-professions}
\end{table}

This active gender debiasing is prominent among professions accessible via the \textit{occupation} property.
We find professions heavily skews towards females, for instance, \textit{educator} and \textit{activist}, 
with even gender-specific professions, such as, 
\textit{actress} (only females) (Figure \ref{fig:gender-ratio-by-profession}).
In particular, GPTKB has 189 professions with no female entities and 203 professions with no male ones (top-10 list in Table \ref{tab:gendered-professions}).
Gender-neutral professions occurring at lower frequencies exhibit skews reflecting traditional gender norms, such as \textit{nurse} with a gender ratio of 39.5 (1,541 triples) and \textit{software\_engineer} with a gender ratio of 0.38 (696 triples). 

\begin{table}[t]
    \centering
    \resizebox{0.85\columnwidth}{!}{%
    \begin{tabular}{l ll | l@{}}
    \toprule
    \textbf{Nationality} &  \textbf{GPTKB} & \textbf{Global}* & \textbf{Global top-10}\\
    \midrule 
    \cellcolor[HTML]{7aefa5}American$^a$ & 31.5 & 4.3 & India (18.2)\\
    British$^{b}$ & 12.4 & 0.8 & China (17.6)\\
    French & 3.9 & 0.8 & United States (4.3)\\
    German & 3.6 & 1.0 & Indonesia (3.5) \\
    \cellcolor[HTML]{7aefa5}Indian & 3.6 & 18.2 & Pakistan (3.9)\\
    Canadian & 3.3 & 0.5 & Nigeria (2.9) \\
    Australian & 2.5 & 0.3 & Brazil (2.6)\\
    Japanese & 2.4 & 1.5 & Bangladesh (2.1)\\
    Italian & 2.3 & 0.7 & Russia (1.8)\\
    \cellcolor[HTML]{7aefa5}Russian & 1.4 & 1.8 & Ethiopia (1.6)\\
    \bottomrule
    \multicolumn{4}{l}{$^a$\small \textit{includes United\_States} $^b$\small \textit{includes English}}
    \end{tabular}
    }
    \caption{Nationality distribution in percentage in GPTKB and the real-world. Nationalities in GPTKB that feature in the global top-10 are highlighted. *Real world data obtained from \url{https://worldpopulationreview.com/countries} as of 2025. 
    }
    \label{tab:nationality}
\end{table}

\begin{table}[t]
    \centering
    \resizebox{\columnwidth}{!}{
    \begin{tabular}{@{}ll@{}}
    \toprule
    \textbf{Nationality} & \textbf{Top-5 Professions} \\
    \midrule
    American & author, politician, actor, actress, musician \\
    
    British & footballer, cricketer, politician, priest, rugby\_union\_player \\
    French & politician, writer, actor, actress, footballer \\
    German & footballer, politician, professor, writer, composer \\
    
    Indian & politician, actor, actress, film\_director, writer \\
    
    Canadian & politician, ice hockey player, actor, actress, musician \\
    Australian & politician, cricketer, actress, actor, footballer \\
    Japanese & footballer, voice\_actor, actor, singer, actress \\
    Italian & footballer, politician, painter, actor, composer \\
    Russian & footballer, writer, politician, actor, military\_officer \\
    \midrule
    \end{tabular}
    }
    \caption{Top professions of most frequent nationalities.}
    \label{tab:professions-by-nationality}
\end{table}

\paragraph{Geographic Bias}
Table \ref{tab:nationality} shows the most frequent nationality distribution in GPTKB. While GPTKB nationalities do not exclusively belong to living humans, we still compare the percentage share with the global population distribution to obtain an estimate of the order of the distribution. The bias to English-speaking nations, making up almost 44\% of all nationalities, is reflective of the training corpus and not alarming. However, in light of the somewhat even gender distribution observed previously, only three countries of the top ten forming global population come up in GPTKB's top ten nationalities. If we exclude American, their share, i.e., Indian and Russian is only 5\%, while Chinese form only 1\%. We find this over-representation of the United States in other geographic distributions as well. 32\% of the 63K cities in GPTKB are located in the United States,
followed by Japan (5.0\%), India (3.2\%), China (2.7\%) and Russia (2.3\%). 22\% of the 73K companies are located in the United States, followed by the United Kingdom (10\%), India (7\%), Japan (5\%) and China (4.5\%).

\paragraph{Composite Biases}
In Table \ref{tab:professions-by-nationality}, we look at the most popular occupations by nationality. Politicians and footballers are by far the most frequent professions across nationalities. When we qualify by gender, politicians have a positive gender skew towards females with female-to-male ratio of 3.12 compared to actors (0.03) and footballers (0.31) (Figure~\ref{fig:gender-ratio-by-profession}).
Figure \ref{fig:gender-ratio-by-nationality} shows the gender skew in the 50 most frequent nationalities. 43\% of the nationalities with a female-to-male ratio >1 belong to English-speaking countries, suggesting selective gender debiasing, as seen for professions.

\paragraph{Political Bias}
We find no strong evidence of a consistent left- or right-wing bias in GPTKB’s factual coverage: U.S. Democrats have more recorded membership triples than that of Republicans (11,781 vs. 9,452), but after normalizing by the number of politician entities, positive facts occur at nearly identical rates (0.359 vs. 0.361 triples per member), while negative facts are only slightly more frequent for Republicans (0.0056 vs. 0.0042 explicit negative triples per member), based on a dictionary of positive (e.g., award) and negative (e.g., convictedOf) relations. For Germany, the corresponding left/right blocs show near parity in positive coverage (0.348 vs. 0.330 triples per member) and virtually no recorded negative facts on either side.

\begin{mybox}{Biases}
    GPTKB exposes a strong English country bias and selective gender debiasing via more female gender assertions and professions in GPT-4.1. The bias flips for non-English speaking countries and gender-neutral professions.
\end{mybox}

\section{Accuracy and Hallucinations}
In this section we examine \textit{RQ3. ``How does factual accuracy change with long-tailed entities?''} We further sample unverifiable triples to detect hallucinations versus long-tailed entities.
\begin{figure}[t]
 \centering
 \begin{subfigure}[b]{0.47\textwidth}
    \includegraphics[width=\columnwidth]{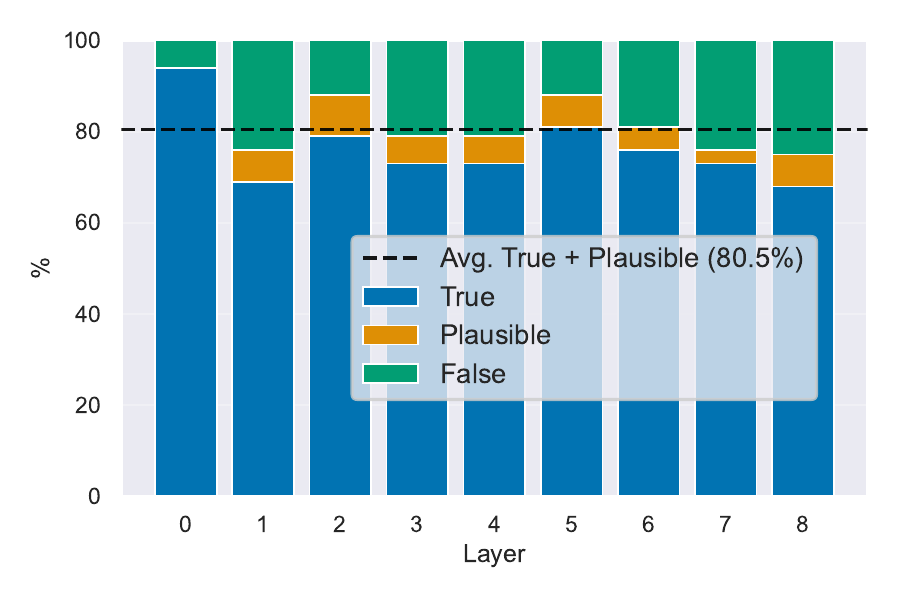}
    \caption{Trend of \textbf{triple accuracy} across layers.
     }
    \label{figure:tripleaccuracy}
 \end{subfigure}
 \hfill
 \begin{subfigure}[b]{0.47\textwidth}
    \includegraphics[width=\columnwidth]{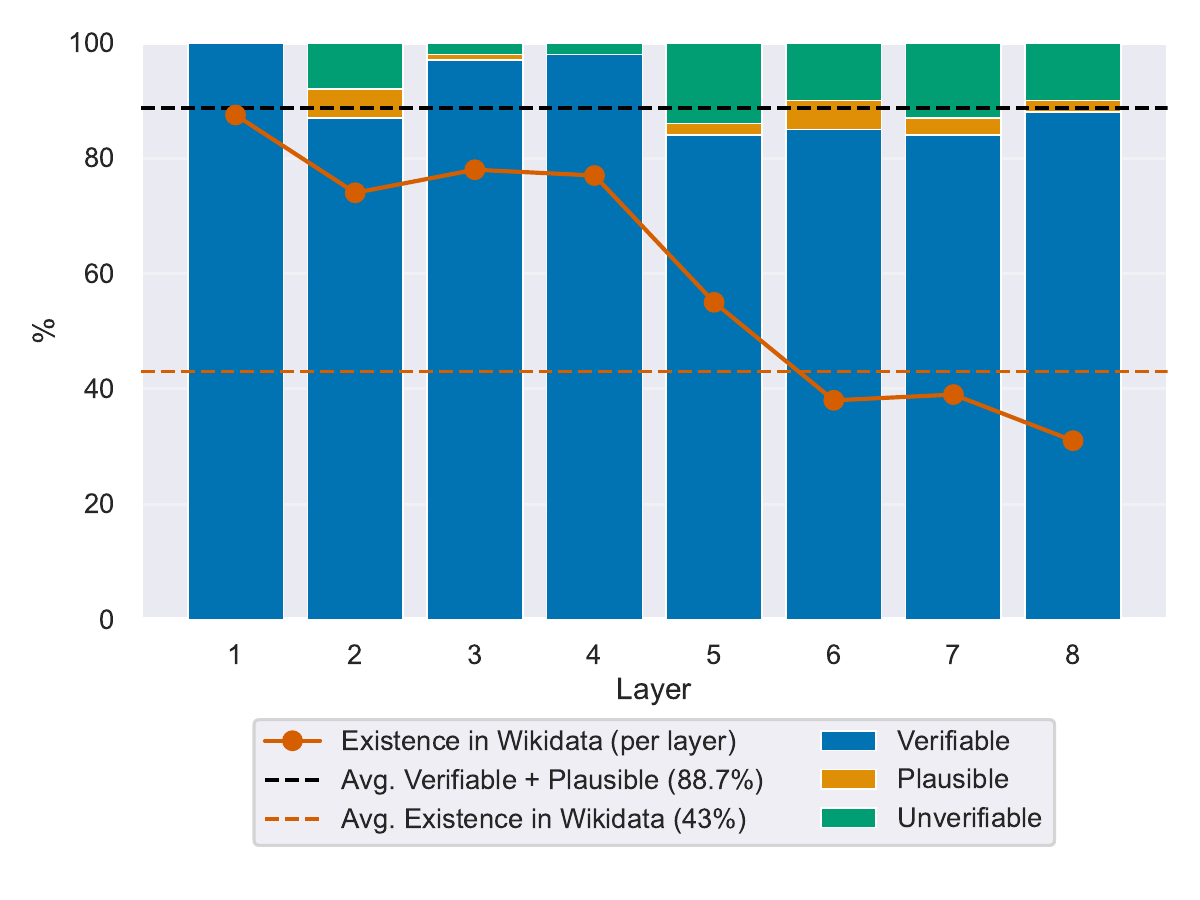}
    \caption{Trend of \textbf{entity verifiability} across layers.
     }
    \label{figure:entityaccuracy}
 \end{subfigure}
 \caption{GPTKB accuracy across layers.}
 \label{figure:accuracy}
 \end{figure}

\paragraph{Accuracy}
GPTKB explores an LLM in a breadth-first manner, such that we get a set of new entities from parent entities of the previous exploration layer. We sampled and validated 100 triples and 100 entities from each layer in the source KB (all triples and entity from seed layer). Following the paradigm in \citet{hu2025enabling}, we use Llama3.1-70B-Instruct as an judge to determine triple accuracy as true/plausible/false when provided with the top-5 webpage snippets for the search term ``$\langle$subject$\rangle$ $\langle$object$\rangle$'' from a search engine API. Similarly, the judge model determines whether an entity is verifiable/plausible/unverifiable when provided with the top-5 search
results for the subject entity.
The results are presented in Figure~\ref{figure:accuracy}.

For factual knowledge particularly relevant is the MMLU benchmark, 
widely used to test broad general knowledge and reasoning across academic disciplines, where
GPT-4.1 is reported to achieve 90.2\%~\cite{openai2025gpt41}.
This starkly contrasts with the factual accuracy of only 75\% and 80.5\% when considering both true and plausible.
Our analysis reveals a notable decoupling between entities' presence in the public domain and LLM's factual accuracy regarding it. Specifically, we observed that as the GPTKB exploration deepened, the percentage of entities verifiable in Wikidata showed a clear downward trend with an average per layer existence in Wikidata at 43\%, while the accuracy of the LLM's elicited statements remained consistent, showing no significant degradation. This strongly indicates that the LLM possesses an internally consistent knowledge base that extends well into the long tail.

\begin{table}[ht]
\centering
\small
\begin{tabular}{lrrr}
\toprule
\textbf{Domain} & \textbf{True} & \textbf{Plausible} & \textbf{False}\\
\midrule
Locations        & 75.7\% & 6.8\% & 17.5\% \\
Persons    & 74.2\% & 6.5\% & 19.3\% \\
Politics   & 72.7\% & 0\% & 27.3\% \\
Organizations  & 72.2\% & 7.8\% & 20.0\% \\
Science        & 70.6\% & 5.6\% & 23.8\% \\
Business     & 68.9\% & 8.1\% & 23.0\% \\
Sports       & 67.5\% & 8.4\% & 24.1\% \\
\bottomrule
\end{tabular}
\caption{Accuracy of triples divided into 7 specific domains.}
\label{tab:accuracy-domain}
\end{table}

When categorized by domain in Table \ref{tab:accuracy-domain},
the LLM's reliability is the highest for locations and persons. Conversely, the LLM performs worst in business and sports domains, where the knowledge is more numeric and time-sensitive. Notably, the politics domain showed a 0\% plausibility rate and the highest error rate (27.3\%), suggesting the LLM mirrors the polarized, binary nature of political discourse, leading to confident but often incorrect assertions.

\paragraph{Hallucinations}
We randomly sampled and manually verified triples from GPTKB until we collected 50 triples that could not be verified as correct.
Of these 32 triples (64\%) had incorrect objects, mostly belonging to related entities, for example, \textit{Ohio's 7th congressional district} - \textit{currentRepresentative} - \textit{William McKinley} is incorrect, though he served the 17th congressional district of Ohio; 
 9 triples (18\%) had either non-existent, ambiguous, or long-tail subjects, for example, \textit{John Stewart (British Army officer, died 1811)} and \textit{Chiara Ricciardone} are both long-tail and ambiguous entities;
 9 triples (18\%) had dubious predicates (likely results of relation clustering errors), for example, \textit{Connections (TV series)} - \textit{hasMethod} has \textit{tt0078588} as its object, which is actually the IMDb ID of the series.

\begin{mybox}{Accuracy}
    GPTKB's accuracy suggest that GPT-4.1's factual accuracy is lower than that suggested by common factuality benchmarks, even though the accuracy does not degrade towards the tail. Most hallucinations occur via incorrect objects on real subjects.
\end{mybox}

\section{Consistency}
In this section and the next we address \textit{RQ4. ``How consistent and temporally coherent is GPTKB?''}

\paragraph{Paraphrased Subjects}  
Predicates denoting aliases, such as, \textit{alsoKnownAs}, \textit{alias} and \textit{realName}, are the most common for persons/humans and fictional characters, occurring for a combined total of 36K triples. 
These alias predicates, occur less frequently in other classes, such as, \textit{city} (1.1K) and \textit{film} (3K). GPTKB contains 112K subjects connected to 124K paraphrased entities via the \textit{alsoKnownAs}, \textit{alias} and \textit{realName} predicates, overcounting at least 2\% of the entities.

\begin{figure}
    \centering
    \includegraphics[width=0.95\columnwidth]{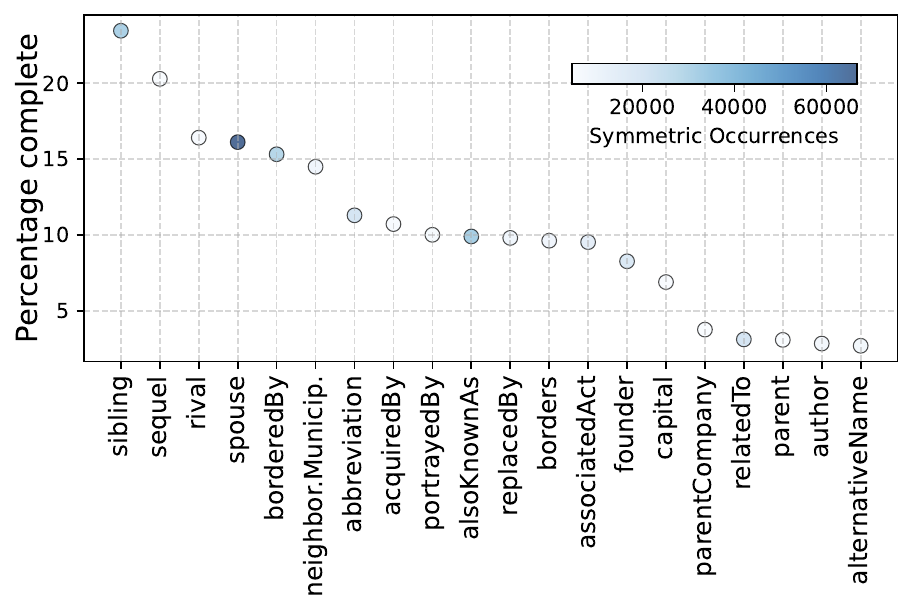}
    \caption{Percentage of total triples present symmetrically for the 20 most frequent symmetric relations.}
    \label{fig:symmetric-relation-completeness}
\end{figure}

\paragraph{Symmetric Relations}
GPTKB has 4,841 relations with at least one reciprocal pair.
Figure \ref{fig:symmetric-relation-completeness} shows the symmetric completeness of 20 most frequent such relations. 
The \textit{sibling} relation is the most complete at 23\%, followed by \textit{spouse} at 16\% completeness. 
We prompted GPT-4.1 for the missing symmetric counterparts of 100 spouse statements to determine whether the missing statements occurred due to incomplete elicitation. The model returned the missing symmetric object only 9\% of the time, an empty response for 13\% of the cases and hallucinated for the rest, showing that the LLM cannot judge relation symmetry. 
We also identify relations, such as, \textit{founder}, \textit{author} and \textit{abbreviation}, that wrongly appear as symmetric relations due to incorrect relation clustering, such as clustering \textit{founder} and \textit{founder of}.

\paragraph{Type Mismatch}
Our analysis surfaces another consolidation challenge via disambiguation errors that lead to type inconsistencies in GPTKB for polysemous named-entities. In family relations, such as, spouse, parent, sibling, children, we notice a shift from fictional characters to humans/persons and vice-versa in more than 4K triples, where the object elicitation resolves to the more prominent entity. For instance, for the spouse of the character Alexander Hamilton, namely Eliza Hamilton, in the play Hamilton is later generated as a person. Type mismatch more indicative of internal LLM inconsistency is for the \textit{children} relation, which has both entities and integer values as objects. 

\begin{mybox}{Consistency}
GPTKB overestimates parametric knowledge due to at least 2\% paraphrased subjects, and symmetric relations are less than 25\% complete. Consolidating paraphrased subjects can improve the coverage of symmetric relations.
\end{mybox}

\begin{figure*}[t]
    \centering
    \begin{subfigure}[t]{0.46\textwidth}
        \centering
        \includegraphics[width=\columnwidth]{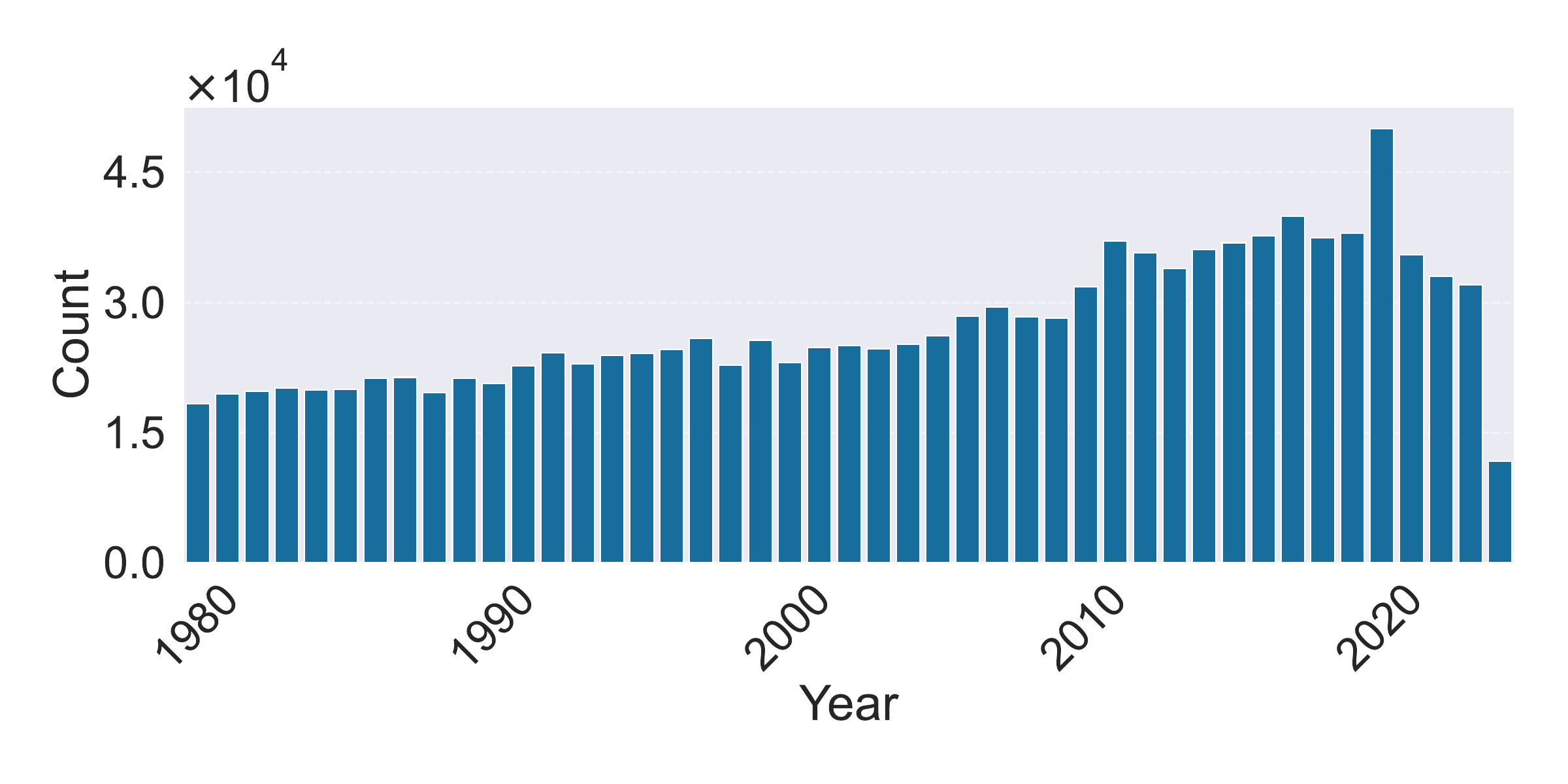}
        \caption{Year distribution with a peak at 2020.}
        \label{fig:v15-year-distribution-1980-2030}
    \end{subfigure}%
    ~ 
    \begin{subfigure}[t]{0.46\textwidth}
        \centering
        \includegraphics[width=\columnwidth]{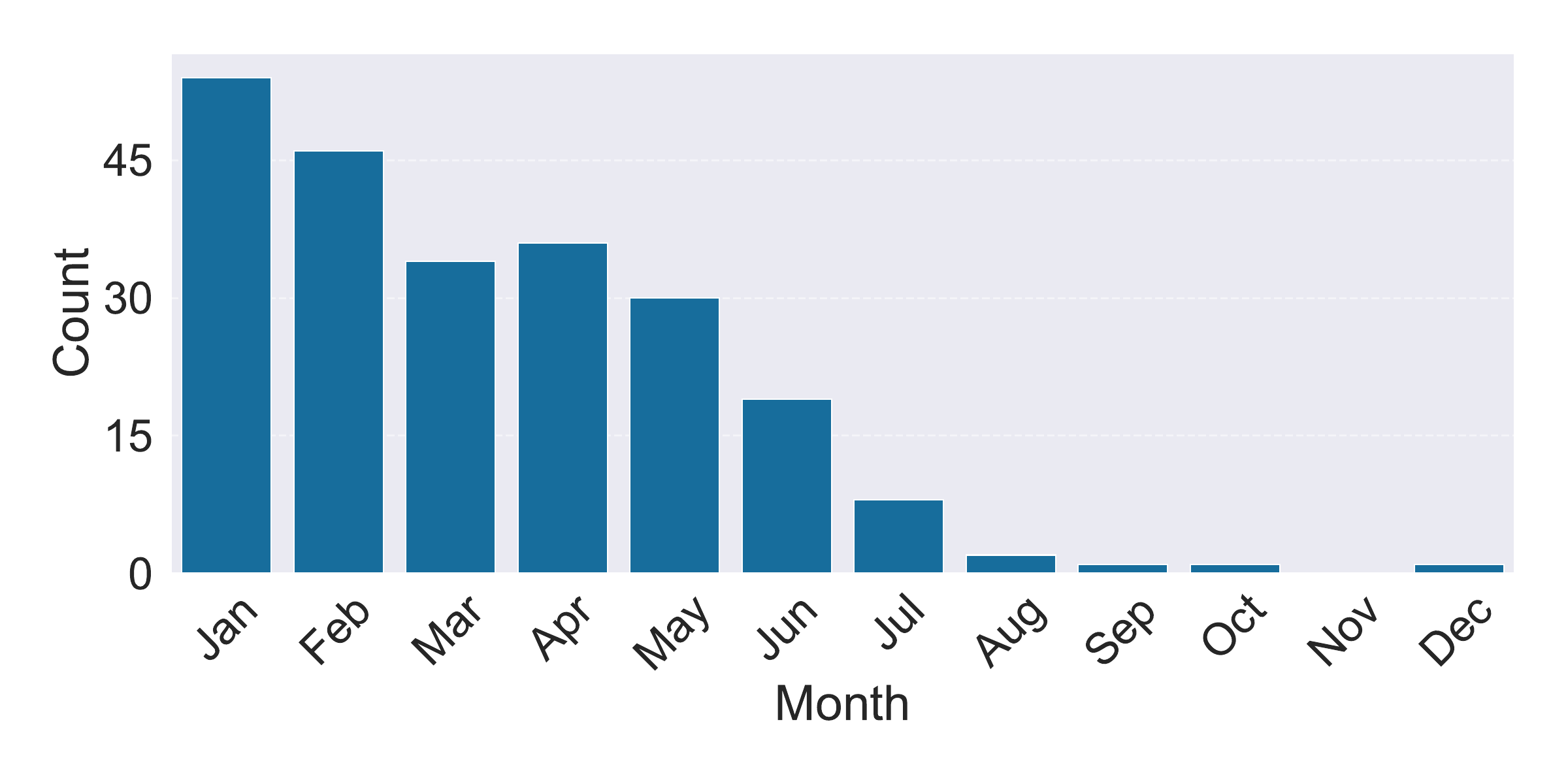}
        \caption{Month distribution of the death dates in 2024.}
        \label{fig:v11-year-distribution-1980-2030}
    \end{subfigure}
    \caption{Timeliness distribution in GPTKB.}
    \label{fig:v15-month-of-death-2024}
\end{figure*}

\begin{figure}[t]
    \centering
    \includegraphics[width=0.95\columnwidth]{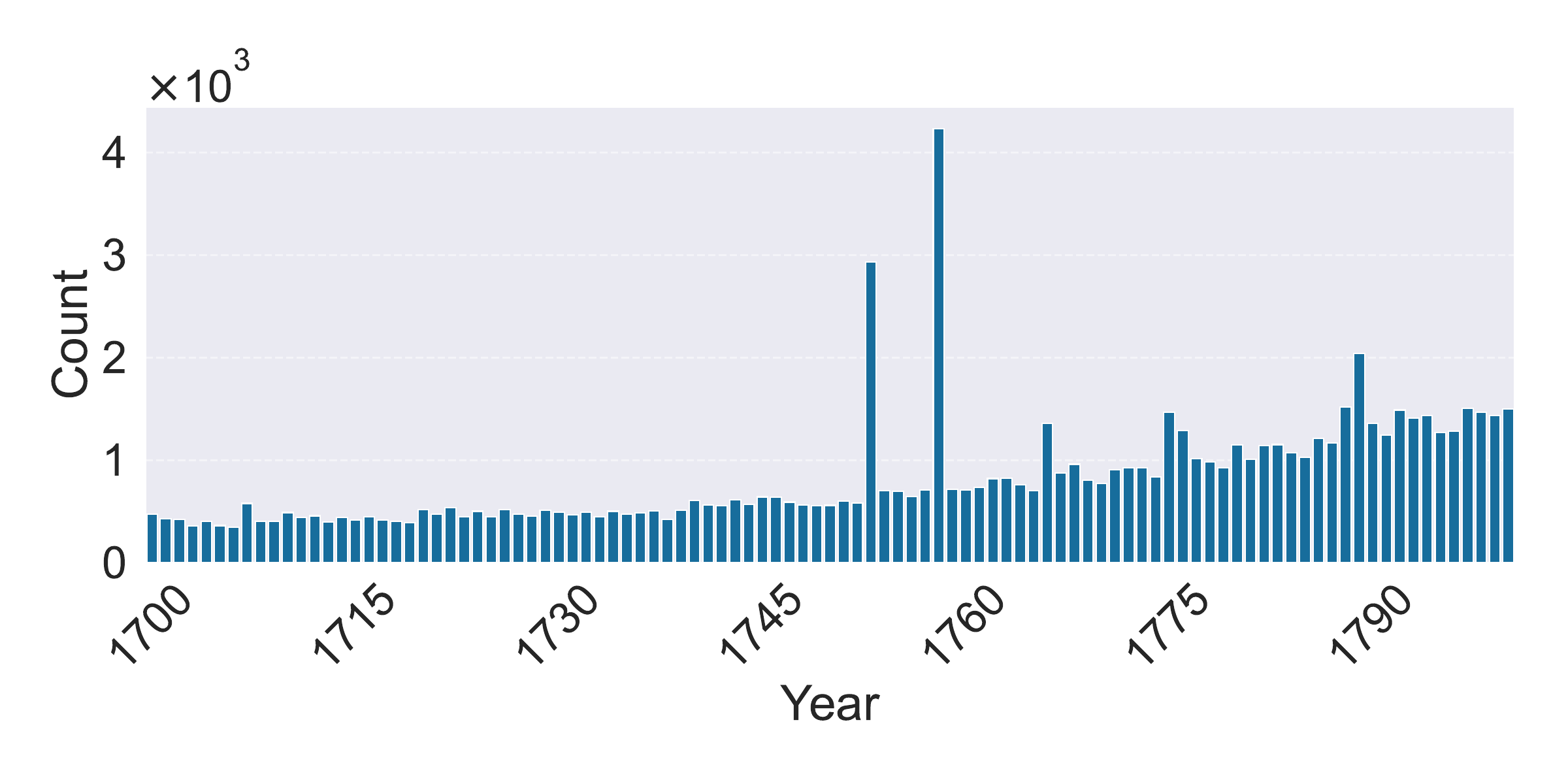}
    \caption{Year distribution (1700-1800) with observed spikes at 1753 and 1758.}
    \label{fig:v15-year-distribution-1700-1800}
\end{figure}

\section{Timeliness}

Timeliness is a crucial aspect of KBs.
LLMs, without access to up-to-date information, are limited to knowledge available up to their training cutoff date.
We assess the timeliness of GPTKB via triples whose objects are timestamps. We find them by filtering triples with predicates that contain either ``year'' or ``date''. The objects are parsed into YYYY-MM-DD format using regular expression. Any missing day or month is substituted with ``00''. This leaves us with 2.7M triples 
whose objects were successfully parsed into the date format.

\paragraph{Latest Observed Years and Recency Bias}

Figure~\ref{fig:v15-month-of-death-2024} (a) presents the year distribution between 1980 and 2030.
There is a pronounced decline in distribution around the reported knowledge cutoff date of GPT-4.1 (June 1, 2024).\footnote{\url{https://platform.openai.com/docs/models/gpt-4.1}}
We zoom into the year 2024, in
figure~\ref{fig:v15-month-of-death-2024} (b), which shows the month distribution of 232 triples that have \emph{deathDate} as their predicate and 2024 as their object year.
The distribution declines drastically in the later months, with less than 20 occurrences in June and the following months.
Furthermore, we observe that GPT-4.1 exhibits a clear recency bias, with more than 75\% of all 2.7M triples having dates after 1917.

\paragraph{Years of Plants and Animals}
Figure~\ref{fig:v15-year-distribution-1700-1800} presents the distribution of years in the 18$^{\text{th}}$ century, highlighting two notable spikes in 1753 and 1758, with 2,934 and 4,233 occurrences, respectively.
In both cases, the most popular predicate is \emph{describedYear} (more than 80\% of all triples), and the subjects of these triples are scientific names of either plants or animals.
Historically, these peaks correspond to pivotal milestones in taxonomy: the publication of Carl Linnaeus's Species Plantarum  \cite{linnaeus1753species} and the 10$^{\text{th}}$ edition of Systema Naturae \cite{linnaeus1758systema}, which laid the foundations of modern botanical and zoological nomenclature.

\begin{mybox}{Timeliness}
    Dates in GPTKB enable a fine-grained tracing of GPT-4.1's knowledge cutoff. The LLM has a recency bias, with significant spikes found in historical editorial events. 
\end{mybox}

\section{Implication for Downstream Tasks}
Our main findings from \textbf{RQ1} are that GPT's knowledge is vast, remembering over 1M persons, but at the same time well-filtered, virtually without any sensitive content. The broad taxonomy versus narrow knowledge content indicates a critical knowledge gap in the underlying LLMs. Also, multi-lingual knowledge clusters indicate complementary language-specific content that remain untapped. 

Our bias quantification via nationality and gender ratios to examine \textbf{RQ2} reveals that the English knowledge is biased towards the English-speaking world and exhibits a bias towards female entities, though not for non-English locales. Given that only three of the global top-10 nationalities make it to the top-10 nationalities in GPTKB and just 22\% of the entities from the classes \textit{human} and \textit{person} mention their gender reveals an imbalanced representation in GPT-4.1 surfaced by GPTKB. 

The stability of the true assertions across layers (Figure~\ref{figure:tripleaccuracy}) and domains (Table~\ref{tab:accuracy-domain}) indicate consistent knowledge that extends into the long tail (\textbf{RQ3}). Yet, this average factual accuracy of 75\% is significantly below other benchmark results. 

Findings from \textbf{RQ4} confirm previous sample-based studies on consistency. We observe significant logical inconsistencies, like relation asymmetry, redundancy, and type mismatches, which can inform future knowledge consolidation strategies. Time-stamped triples help estimate the knowledge cut-off indicating where newer information is necessary. GPTKB preserves historical editorial events evident from significant spikes. 

\section{Conclusion}
With this paper we provide the framework and the first multi-dimensional analysis of LLM knowledge at scale, based at over 100M facts extracted from GPT-4.1 into GPTKB v1.5 \cite{GPTKB15}. Due to cost, this analysis cannot be easily repeated with other LLMs, but our results stand exemplarily for the knowledge of a 2025 frontier model, and notably, could be performed even for closed-source API-only models.

\section{Limitations}
In this work, we analyze a closed-source, commercial model. Without access to the model's training data, we cannot estimate the LLM's capacity for retaining knowledge. Access to training data, as in the case of open-source models, would provide a deeper insight into knowledge coverage and bias propagation from the data to the LLM parameters and generated knowledge. Nonetheless, the presented analysis paradigm stands out for enabling to gain insights into potentially any LLM, including closed source, commercial ones, as we indeed show in this work. This might help us to shed light on such models' training data and post-training practices e.g., active gender de-biasing. 

While our analysis offers empirical evidence regarding the quality of knowledge stored in GPT-4.1—examining aspects such as accuracy, bias, and inconsistencies, as well as its similarities and differences compared to classical knowledge sources like Wikidata and the Web—these findings may not necessarily generalize to other LLMs.

\section{Ethical Considerations}
When analyzing very large amounts of data, such as GPTKB, there is a potential risk of encountering sensitive information. However, our findings indicate that such instances are absent in GPTKB. It is important to note that our analysis does not offer formal privacy guarantees and should be distinguished from research specifically focused on privacy leakage in LLMs under extraction attacks \cite{brown2022what,lukas2023analyzing}.

\bibliography{custom}

\end{document}